\documentclass[times,twocolumn,final]{elsarticle}

\usepackage{ipcai_arxiv}
\usepackage{framed,multirow}

\usepackage{amssymb}
\usepackage{latexsym}

\usepackage{url}
\usepackage{cite}
\usepackage{amsmath,amssymb,amsfonts,bm}
\usepackage{booktabs}
\usepackage{arydshln}
\usepackage{tabulary}
\usepackage{graphicx,epstopdf}
\usepackage{textcomp}
\usepackage{subcaption}
\usepackage{booktabs}
\usepackage[svgnames,table]{xcolor}
\usepackage{pifont}
\usepackage{arydshln}
\usepackage[colorlinks=true,citecolor=cyan,urlcolor=cyan]{hyperref}
\usepackage{graphicx}
\usepackage{placeins}
\usepackage{fancyhdr}
\usepackage{float}

\usepackage{listings}%
\usepackage{lipsum}
\usepackage{makecell}

\newcommand{\xmark}{\ding{55}}%

\usepackage{xspace}

\newcommand{\etal}{\textit{et al.}\xspace}

\begin{document}


\title{Self-Supervised Multi-View 3D Gaze Target Estimation via Probabilistic Ray Marching}

\author[1,2]{Keqi \snm{Chen}\corref{cor}}

\author[1,2,3]{Vinkle \snm{Srivastav}}

\author[1,2]{Nicolas \snm{Padoy}}

\cortext[cor]{Corresponding author: keqi.chen@unistra.fr}

\address[1]{University of Strasbourg, CNRS, INSERM, ICube, UMR7357, France}

\address[2]{IHU Strasbourg, 67000 Strasbourg, France}

\address[3]{Department of Data Science and AI, Wadhwani School of Data Science and AI (WSAI), Indian Institute of Technology (IIT) Madras, 600036 Chennai, India}

\received{XXX}
\finalform{XXX}
\accepted{XXX}
\availableonline{XXX}
\communicated{XXX}

\begin{abstract}
We present a self-supervised approach, \emph{Self-MVGTE}, for estimating 3D gaze targets from multiple camera views. Unlike existing methods that independently estimate 2D gaze targets per camera view, Self-MVGTE predicts gaze targets directly in 3D space for the first time. Moreover, it does not require any ground-truth annotations from the target scene and uses only the multi-view input images from a calibrated camera setup, pseudo 2D gaze target labels from a monocular gaze target estimation model, and 3D gaze vectors from a monocular 3D gaze estimation model. A key challenge is that these pseudo labels are inherently noisy and multi-view inconsistent. To address this, we propose a probabilistic ray marching framework, which models the uncertainty of these pseudo labels and exploits 3D gaze vectors as geometric priors. Specifically, these gaze vectors are first integrated into the monocular gaze target estimation model to improve its generalization to unseen scenes, producing higher-quality pseudo labels. Then, for 3D gaze target estimation, we construct a 3D gaze cone by casting a bundle of rays from the eye position around the gaze vector to strictly constrain the solution space. Within this cone, we propose a depth-guided feature sampling strategy using off-the-shelf DINOv2 and Depth-Anything-3 models, and estimate a spatial likelihood distribution of the gaze target. Finally, we convert the pseudo gaze target labels into a target distribution and softly optimize the network. Extensive experiments on the MVGT dataset show that Self-MVGTE achieves state-of-the-art performance, surpassing existing fully-supervised baselines.
\end{abstract}

\maketitle

\section{Introduction}
\label{sec:intro}

Gaze is an important component of daily human interaction with the world, serving as a critical cue for interpreting human intent, social dynamics, and human-object interactions~\citep{emery2000eyes,fathi2012learning,huang2015using,lukander2017inferring,wei2018and}. To infer where individuals are looking, gaze-following, also known as gaze target estimation, has emerged as an important research field in computer vision. However, existing work mainly concentrates on monocular gaze target estimation, which is inherently constrained by single-view ambiguity. When the human face is not clearly visible or the scene is highly complex, precisely localizing the gaze target from a single perspective is extremely difficult. Furthermore, gaze targets can frequently reside outside the camera frame in monocular setups, hindering consistent gaze target estimation. 

To address these limitations, multi-view gaze target estimation has recently been introduced~\citep{miao2025multi}, which uses multi-camera setups to provide multiple perspectives of both human faces and the surrounding environment. This pioneering work constructed the first Multi-View Gaze Target (MVGT) dataset and proposed a baseline model (hereafter referred to as MV-GTE), demonstrating substantial performance improvements over monocular approaches with the aid of complementary camera views. Nevertheless, training MV-GTE requires ground-truth 2D gaze target labels in each view, which are labor-intensive to acquire.
Furthermore, MV-GTE still formulates the task as a per-view 2D prediction problem, as the monocular models do, instead of predicting a unified 3D coordinate in 3D space. Consequently, it complicates the inference by requiring repeated predictions for each view and inherently risks potential violation of multi-view geometric constraints. Therefore, how to address multi-view gaze target estimation in 3D space and how to eliminate the reliance on labor-intensive annotations remain unresolved challenges.

\begin{figure}[t!]
    \centering
    \includegraphics[width=1.0\linewidth]{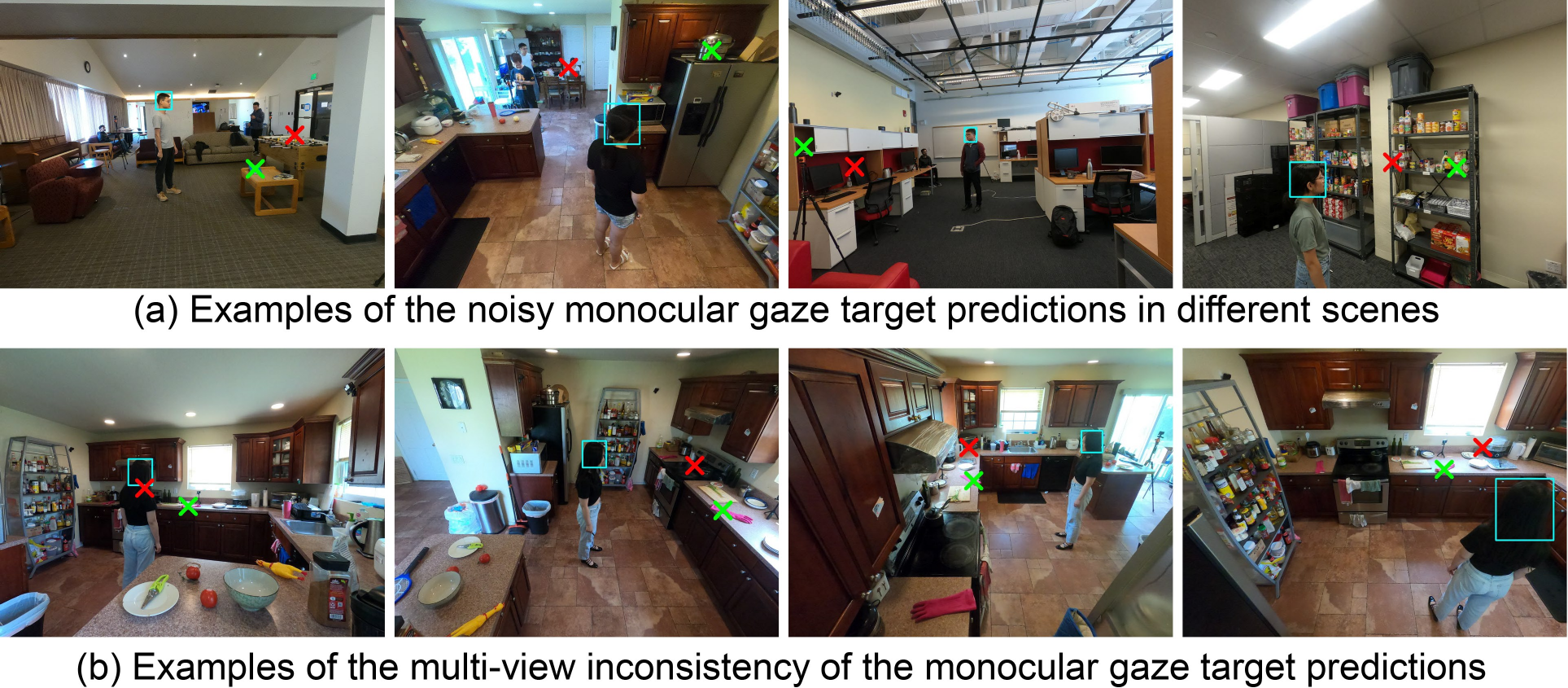} 
    \caption{Comparison between ground-truth gaze target labels (green) and the monocular gaze target predictions (red) obtained from Gaze-LLE~\citep{ryan2025gaze}.}
    \label{fig:intro}
\end{figure}

Inspired by the learning-by-projection paradigm~\citep{chen2019learning} and its successful applications in self-supervised multi-view 3D human pose estimation~\citep{kocabas2019self,srivastav2024selfpose3d}, we propose to train a multi-view 3D gaze target estimation model by comparing the 2D projections of its 3D predictions against the 2D pseudo labels generated from an off-the-shelf monocular gaze target estimation model. However, compared to human pose, monocular pseudo gaze target labels are inherently less reliable for two reasons: (1) they can easily drift between objects under facial occlusion or within complex scenes, which introduces severe noise in the 2D image plane, as shown in Figure~\ref{fig:intro}a; (2) they can be highly inconsistent across different views, violating epipolar constraints, as shown in Figure~\ref{fig:intro}b. Consequently, forcing a model to strictly regress to these erroneous pseudo labels risks not only overfitting to the noise but also producing severely erroneous solutions in 3D space. To overcome these challenges, we propose to constrain the 3D search space using geometric priors~\citep{hartley2003multiple} and adopt probabilistic modeling~\citep{kendall2017uncertainties} to model the uncertainty of the pseudo labels.

In this work, we propose \emph{Self-MVGTE}, a self-supervised learning-based approach that conducts multi-view 3D gaze target estimation from a few calibrated cameras without using any ground-truth annotations from the target scene. Specifically, our approach relies on pseudo 2D gaze target labels obtained from a pretrained monocular gaze target estimation model~\citep{ryan2025gaze}. To address the geometric ambiguity and noise in these pseudo labels, we use estimated 3D gaze directions as geometric priors to improve their quality and introduce a probabilistic ray marching strategy that constrains the 3D search space while modeling pseudo label uncertainty via probabilistic learning. 

To prepare 2D gaze target labels for training, we adopt the state-of-the-art Gaze-LLE model~\citep{ryan2025gaze}, which leverages a frozen DINOv2 encoder~\citep{oquab2023dinov2} and a lightweight decoder. 
Although Gaze-LLE demonstrates strong cross-dataset generalization in unseen scenes, we hypothesize that incorporating estimated 3D gaze vectors as geometric priors can further improve its performance. This is because 3D gaze estimation, which relies exclusively on facial features, generalizes better to unseen scenes than 2D gaze target estimation. 
Therefore, we use an off-the-shelf monocular 3D gaze estimation model~\citep{vuillecard2025enhancing} trained on the Gaze360~\citep{kellnhofer2019gaze360} and GazeFollow~\citep{recasens2015they} datasets to predict a 3D gaze vector for each person. 
We encode these gaze vectors as additional input tokens to the decoder and retrain Gaze-LLE on the GazeFollow~\citep{recasens2015they} and VideoAttentionTarget~\citep{chong2020detecting} datasets. 
The resulting model is then applied to the multi-view dataset to estimate 2D gaze targets in each view. Finally, we filter out predictions for which the angular deviation between the projected 2D gaze vector and the eye-to-target vector exceeds a specified threshold. The remaining predictions are used as pseudo labels for training.

Unlike existing monocular or multi-view gaze target estimation approaches that predict a 2D gaze target in each view, we propose to estimate the gaze target directly in 3D space. However, searching an unconstrained 3D space for a surface point without ground-truth labels or accurate metric depth is highly ill-posed. To explicitly constrain the search space and model label uncertainty, we propose a probabilistic ray marching strategy that uses 3D gaze vectors as geometric priors. Specifically, we first locate the 3D eye position through eye detection and triangulation, and then compute a multi-view 3D gaze vector by averaging the monocular predictions. Afterwards, starting from the eye position, we construct a 3D gaze cone by casting a bundle of rays around the gaze vector.
We then sample a set of query points along each ray and aggregate their corresponding multi-view features to obtain point-level representations. These representations are used to estimate a spatial likelihood distribution of the 3D gaze target over the entire 3D gaze cone. 

To compute point-level representations for 3D gaze target localization, we need both semantic information for identifying potential target objects and depth information for reasoning about visibility along the line of sight. To this end, following Gaze-LLE~\citep{ryan2025gaze}, we first use a frozen DINOv2~\citep{oquab2023dinov2} encoder to extract semantic scene features for each image. Then, we introduce a soft depth-guided visual attention mechanism. Specifically, we estimate multi-view metric depth using Depth-Anything-3 (DA3)~\citep{lin2025depth} and measure the distance between the depth of each projected query point and the corresponding DA3 surface. These distances serve as soft geometric priors to suppress semantic features from views in which the query point is occluded. We additionally encode these distances using a CNN to obtain depth features. The semantic and depth features are then passed through a decoder to obtain point-level representations. 
After predicting logit scores and applying a global softmax, we obtain a spatial likelihood distribution over the 3D gaze cone and compute its expectation as the final gaze target location. 
To further determine whether the gaze target lies within each 2D view, we project the estimated 3D gaze target onto each view and compute its distances to the image boundaries. This naturally serves as a differentiable geometric gate. Combining this gate with learned semantic and depth features, we obtain more robust in-or-out predictions.

Finally, we train the multi-view model using the prepared pseudo 2D gaze target labels. Instead of directly supervising the predicted 3D gaze target with noisy 2D labels, we convert these pseudo labels into a 3D target distribution to supervise the estimated spatial likelihood distribution. Specifically, for each query point, we compute its 2D distance to the pseudo labels and its depth distance to the nearest DA3 surface, which provide semantic- and depth-based probabilities, respectively. By applying softmax over all query points within the 3D gaze cone, we obtain the target distribution. We then minimize the Kullback-Leibler (KL) divergence between the target distribution and the estimated spatial likelihood distribution. For in-or-out prediction, we directly apply a cross-entropy loss using the pseudo in-or-out labels, because a geometric gate based on 3D-to-2D projection is already applied as mentioned above. 

We evaluate our approach on the MVGT dataset~\citep{miao2025multi} and conduct extensive ablation studies. Following existing baselines~\citep{miao2025multi}, we perform leave-one-scene-out cross-validation. Experimental results show that our approach outperforms existing fully-supervised approaches.

We summarize our contributions as follows: (1) We improve the generalization of vision foundation model-based monocular gaze target estimation to unseen scenes by integrating estimated 3D gaze vectors as geometric priors. (2) We address multi-view 3D gaze target estimation for the first time by directly localizing gaze targets in 3D space. (3) We propose a self-supervised multi-view 3D gaze target estimation approach that requires no manual annotations from the target scenes and surpasses fully-supervised baselines. (4) We introduce a probabilistic ray marching strategy that leverages 3D gaze vectors as geometric priors to constrain the 3D search space and addresses 3D gaze target estimation as a probabilistic learning task using noisy pseudo labels.

\section{Related work}
\label{sec:related}

In this section, we briefly review current works related to gaze target estimation, unconstrained 3D gaze estimation, self-supervised learning-by-projection, and probabilistic ray marching.

\noindent\textbf{Gaze target estimation:} The gaze target estimation task was first introduced by Recasens~\etal~\citep{recasens2015they}, where they contribute the GazeFollow dataset and propose a two-branch network that separately estimates scene saliency and gaze direction. Subsequently, several following works adopte this multi-branch design and incorporate additional cues, such as object detection~\citep{saran2018human,guan2019enhanced,hu2022gaze,wang2022gatector}, human pose estimation~\citep{guan2019enhanced,gupta2022modular}, depth~\citep{fang2021dual,bao2022escnet,jin2022depth,gupta2022modular,tonini2022multimodal,miao2023patch,tafasca2023childplay,horanyi2023they}, 2D/3D gaze direction~\citep{chong2017detecting,chong2018connecting,saran2018human,lian2018believe,fang2021dual,jin2022depth,gupta2022modular,tafasca2023childplay,horanyi2023they,tafasca2024sharingan}, eye detection~\citep{fang2021dual}, etc. A few recent works propose to use transformer architecture to jointly locate heads and predict gaze targets in an end-to-end way~\citep{tu2022end,tonini2023object,tu2023joint}, but these models require complex training. In order to both simplify the multi-branch pipeline and improve adaptability, Ryan~\etal~\citep{ryan2025gaze} propose Gaze-LLE, which marries gaze target estimation with large vision foundation models. By leveraging a frozen DINOv2 encoder~\citep{oquab2023dinov2} and a lightweight decoder, Gaze-LLE significantly reduces trainable parameters, achieves state-of-the-art performance on several benchmarks, and demonstrates strong generalization ability. In this work, we propose to further enhance the generalization of Gaze-LLE using estimated 3D gaze directions as geometric priors. 

Despite these advances in monocular gaze target estimation, it is inherently constrained by single-view ambiguity, especially when the human face is not clearly visible or the scene is highly complex. Therefore, Miao~\etal~\citep{miao2025multi} introduce the task of multi-view gaze target estimation and contribute the MVGT dataset as a challenging benchmark with four different scenes. They also propose a strong baseline model that effectively aggregates multi-view gaze and scene information and surpasses monocular models by a large margin. However, their method relies on ground-truth labels for training and is still limited to per-view 2D gaze target estimation, as the monocular models do. In contrast, our work explores multi-view 3D gaze target estimation for the first time without using ground-truth labels. 

\noindent\textbf{Unconstrained 3D gaze estimation:} To solve unconstrained 3D gaze estimation without constraint on the head pose, Kellnhofer~\etal~\citep{kellnhofer2019gaze360} introduce the first large-scale dataset Gaze360, and propose an appearance-based temporal model. Subsequently, several works propose to learn multi-scale head, face, and eye features~\citep{ashesh2021gaze360,guan2023end} for better within-data performance. In order to improve generalization to unseen scenes, recent studies have increasingly leveraged weak supervision from more diverse datasets. 
Kothari~\etal~\citep{kothari2021weakly} exploits Looking At Each Other (LAEO) labels and the associated geometric constraints to provide weak 3D gaze supervision from the LAEO datasets. Vuillecard and Odobez~\citep{vuillecard2025enhancing} convert 2D gaze target labels to pseudo 3D gaze labels with broader gaze distributions, and propose GaT to improve both within-domain and cross-domain performance. In this work, we use GaT trained on GazeFollow~\citep{recasens2015they} and Gaze360~\citep{kellnhofer2019gaze360} to estimate 3D gaze directions as geometric priors in an unseen dataset. 

\noindent\textbf{Self-supervised learning-by-projection:} The learning-by-projection paradigm addresses the scarcity of ground-truth data by projecting 2D/3D predictions onto 2D image planes and comparing them with 2D observations. It has been widely adopted in 3D object reconstruction~\citep{chen2019learning,wu2020unsupervised,li2020self}, human pose estimation~\citep{kocabas2019self,srivastav2024selfpose3d}, depth estimation~\citep{godard2019digging,moon2024ground,lavreniuk2025spidepth}, and novel view synthesis~\citep{mildenhall2021nerf,kerbl20233d}. In this work, we apply the learning-by-projection paradigm on multi-view 3D gaze target estimation for the first time using 2D pseudo labels generated from monocular gaze models. 

\noindent\textbf{Probabilistic ray marching:} Originating from volumetric rendering~\citep{max1995optical}, ray marching has become a cornerstone of modern 3D vision, largely popularized by Neural Radiance Fields (NeRF)~\citep{mildenhall2021nerf}, which synthesizes novel views by querying 5D coordinates along camera rays. Subsequent works have introduced probabilistic formulations to model uncertainty in NeRF. S-NeRF~\citep{shen2021stochastic} models a probability distribution over possible radiance fields using variational inference, while CF-NeRF~\citep{shen2022conditional} further models such distributions using conditional normalizing flows.
In this work, we also formalize 3D gaze target estimation as a probabilistic ray marching problem by projecting rays around the gaze direction and predicting a spatial likelihood distribution. 

\section{Methodology}
\label{sec:method}

\subsection{Problem overview}

Given a training dataset of multi-view images $\mathcal{D} = \left\{ \mathbf{I}| \mathbf{h} \right\}$ where $ \mathbf{I} \in \mathbb{R}^{C \times 3 \times H \times W}$ is a multi-view image set from $C$ cameras with height $H$ and width $W$, and $\mathbf{h} \in \mathbb{R}^{C \times 4}$ represents detected head bounding boxes, the goal is to learn a model that estimates the 3D gaze target $\mathbf{X}_{\text{gaze}} \in \mathbb{R}^{3}$ from the multi-view input image $\mathbf{I}$, and predicts an in-or-out probability score $\mathbf{p}_{\text{in/out}} \in [0, 1]$ that represents whether the gaze target is inside each camera view. It should be noted that, in some views, the head bounding boxes may not be detected due to occlusion or limited camera perspective. 

\begin{figure*}
    \centering
    \includegraphics[width=0.9\textwidth]{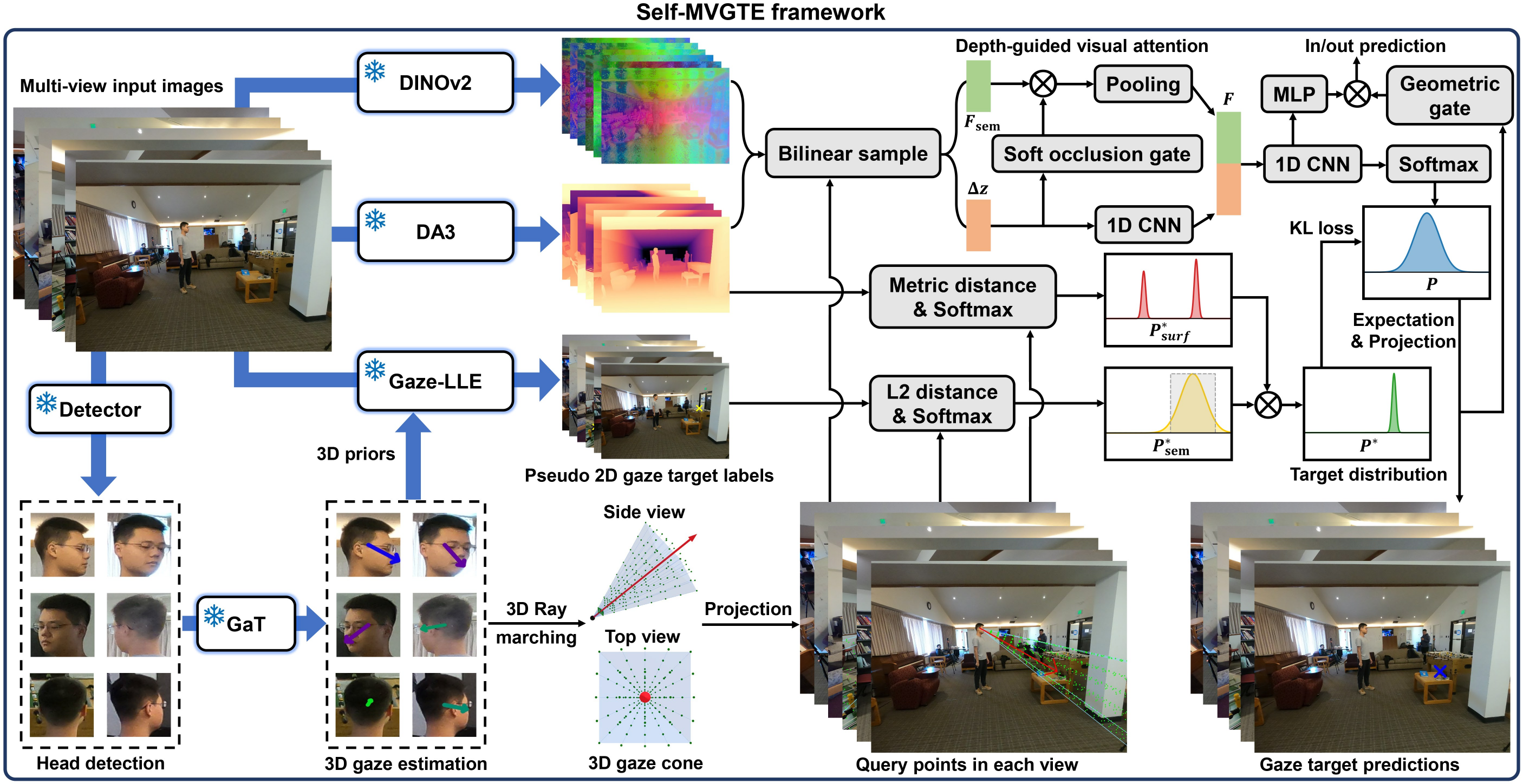} 
    \caption{Framework of our self-supervised approach. Given multi-view input images, we detect heads~\citep{miao2025multi}, estimate 3D gaze vectors~\citep{vuillecard2025enhancing}, and generate pseudo 2D gaze target labels using Gaze-LLE~\citep{ryan2025gaze} with gaze vectors as 3D priors. We extract semantic features using DINOv2~\citep{oquab2023dinov2} and estimate metric depth using Depth-Anything-3 (DA3)~\citep{lin2025depth}. We construct a 3D gaze cone around the averaged gaze vector, sample several query points, project them to each view for semantic and depth feature aggregation, and estimate a spatial likelihood distribution. By converting the pseudo labels into a target distribution, we train the model using the KL loss and localize the gaze target by computing the expectation. Using the gaze target location as a differentiable geometric gate, we predict the in-or-out score in each view. }
    \label{fig:method}
\end{figure*}

In the following, we present our self-supervised learning framework as shown in Figure~\ref{fig:method}. We first improve Gaze-LLE~\citep{ryan2025gaze} using 3D gaze vectors as geometric priors for monocular gaze target estimation and generate pseudo 2D gaze target labels. Then, we propose a ray marching strategy that constructs a 3D gaze cone around the estimated gaze direction as a constrained search space. Within this gaze cone, we apply depth-guided multi-view feature sampling to construct point-level feature representation. Finally, we conduct probabilistic 3D gaze target localization and train the model through probabilistic learning. 

\subsection{Monocular gaze target estimation via 3D priors}

To generate pseudo 2D gaze target labels in unseen scenes, we need a monocular gaze target estimation model with strong cross-dataset generalization. Specifically, we improve Gaze-LLE~\citep{ryan2025gaze} using 3D gaze vectors as geometric priors. Gaze-LLE learns a gaze decoder with a novel head-prompting design on top of a frozen DINOv2~\citep{oquab2023dinov2} encoder, which leverages large-scale pretrained vision knowledge. Building on this architecture, we additionally use an off-the-shelf 3D gaze estimation model GaT~\citep{vuillecard2025enhancing} to predict 3D gaze vectors, which are passed to the gaze decoder together with the DINOv2 features.

Specifically, given an RGB image of size $H \times W$, we pass it through DINOv2 and a learnable linear layer to obtain a lower resolution feature map $D \in \mathbb{R}^{d_{\text{model}} \times H_D \times W_D}$, where $d_{\text{model}}$ is the token size of the transformer gaze decoder. Then, based on the head position, we construct a downsampled, binarized mask $M$ of size $H_D \times W_D$. By learning a head position embedding $e_{\text{head}} \in \mathbb{R}^{d_{\text{model}}}$, we obtain the scene feature map $S$: 
\begin{equation}
    S = D + (M * e_{\text{head}}).
\end{equation}

To encode 3D priors, we pass the cropped head image through GaT and obtain the predicted 3D gaze vector $V_g \in \mathbb{R}^{3}$. It is then projected into gaze token $t_g \in \mathbb{R}^{d_{\text{model}}}$ using a two-layer MLP. Afterwards, we flatten $S$ into a scene token list $[s_1, s_2,...,s_{H_D\times W_D}]$ and concatenate it with gaze token $t_g$ and a learnable task token $t_{\text{in/out}}$ as follows: 
\begin{equation}
T=[t_{\text{in/out}},{t_g},{s_1, s_2,...,s_{H_D \times W_D}}]. 
\end{equation}

Then, we pass $T$ through a three-layer transformer encoder and obtain updated $t_{\text{in/out}}'$ and $S'$. By passing $S'$ through two convolutional layers and up-sampling the output to size $H \times W$, we obtain a classification score for each pixel as being the gaze target or not. By passing $t_{\text{in/out}}'$ through a two-layer MLP, we obtain a score of whether the gaze target is inside the image. We use a binary cross-entropy loss to supervise the training on the GazeFollow~\citep{recasens2015they} and VideoAttentionTarget~\citep{chong2020detecting} datasets. 

Finally, we apply the trained model to the multi-view dataset to predict the 2D gaze target and in-or-out score in each view. Using $V_g$ as 3D priors, we explicitly filter out erroneous predictions by projecting $V_g$ onto the 2D image planes and calculating the angular deviations with the predicted 2D gaze vectors. If the angular deviation exceeds $\theta_{\text{err}}$, we consider the prediction unreliable. After filtering, we obtain the final pseudo labels for the 2D gaze target $\textbf{X}_{\text{gaze}}^*$ and in-or-out score $\textbf{p}_{\text{in/out}}^*$ in the multi-view dataset. 

\subsection{Multi-view 3D ray marching}

In order to locate the 3D gaze target, instead of searching an unconstrained 3D space, we propose a multi-view ray marching strategy, where we search a 3D gaze cone constructed around the estimated 3D gaze direction. 

Given multi-view images where a person's head is visible in more than one view, we first detect the 2D eye positions and calculate the mid-eye point. When eye detection fails, we approximate the mid-eye point from the head bounding box. We then use triangulation to calculate the 3D mid-eye position $X_{\text{eye}} \in \mathbb{R}^{3}$. Next, we use GaT~\citep{vuillecard2025enhancing} to estimate a normalized 3D gaze vector for each visible head. These gaze vectors of different views are averaged and normalized to obtain the multi-view gaze vector $V_{g} \in \mathbb{R}^{3}$. Therefore, we parameterize the 3D gaze ray as follows:
\begin{equation}
X_{d, V_{g}} = X_{\text{eye}} + d \cdot V_{g}, 
\end{equation}
where $d$ denotes the distance along the gaze ray.

To construct a 3D gaze cone, we generate a bundle of rays surrounding $V_g$. Specifically, we define a virtual 2D tangent plane at unit distance from $X_{\text{eye}}$ along $V_g$. We construct a local orthogonal basis on this plane using a global reference vector $V_{\text{world}}=[0,1,0]^T$:
\begin{equation}
    V_{\text{right}} = \frac{V_{g} \times V_{\text{world}}}{\|V_{g} \times V_{\text{world}}\|_2}, \quad V_{\text{up}} = \frac{V_{\text{right}} \times V_{g}}{\|V_{\text{right}} \times V_{g}\|_2}. 
\end{equation}
Given a maximum spread angle $\theta_{\text{ang}}$, the corresponding maximum offset on the unit-distance tangent plane is $\tan\theta_{\text{ang}}$. We uniformly sample an $N\times N$ grid of planar coordinates $(\alpha_j,\beta_j)$ within $[-\tan\theta_{\text{ang}},\tan\theta_{\text{ang}}]$.  
By adding these planar offsets to $V_{g}$ and re-normalizing, we generate a bundle of $N^2$ distinct gaze rays. The direction vector of the $j$-th ray in the bundle is defined as:
\begin{equation}
    V_j = \frac{V_{g} + \alpha_j V_{\text{right}} + \beta_j V_{\text{up}}}{\|V_{g} + \alpha_j V_{\text{right}} + \beta_j V_{\text{up}}\|_2}, \quad \forall j \in \{1, \dots, N^2\}. 
\end{equation}

Afterwards, we dynamically compute the metric depth bounds $d \in [d_{\text{min}}, d_{\text{max}}]$ by calculating the intersection of the central ray $V_{g}$ with the multi-view camera frustums. Given a camera $c$ with intrinsic matrix $K^{c}$ and extrinsic matrix $[R^{c} | T^{c}]$, a point $X_{d, V_j}$ along $V_j$ is projected onto the homogeneous 2D image plane as:
\begin{equation}
\begin{aligned}
[x^{c}_{\text{homo}}, y^{c}_{\text{homo}}, z^{c}_{\text{homo}}]^{\mathrm{T}}
= K^c (R^c X_{d, V_j} + T^c)\\
= d K^c R^c V_j + K^c (R^c X_{eye} + T^c)
\end{aligned}
\end{equation}
A 3D point lies within the visible image frame $(W, H)$ if it satisfies five linear inequalities: it must be in front of the camera ($z_{\text{homo}} > 0$), and its 2D coordinates must satisfy ($0 \le x_{\text{homo}} \le W z_{\text{homo}}$) and ($0 \le y_{\text{homo}} \le H z_{\text{homo}}$). Therefore, we calculate $[d^{c}_{\text{min}}, d^{c}_{\text{max}}]$ for each camera, and then obtain the global bounds of the search space, where $\tau_{\text{head}}$ is a predefined margin that prevents sampling inside the head:
\begin{equation}
    d_{\text{min}} = \max(\tau_{\text{head}}, \min_{c \in [1,...,C]} d_{\text{min}}^{c}), \quad d_{\text{max}} = \max_{c \in [1,...,C]} d_{\text{max}}^{c}.
\end{equation}

Within these global bounds, we uniformly sample $B$ query points along each ray to estimate the gaze target probability during inference. During training, we add uniform random jitter to the sampled distances along the ray to enable the network to learn continuous spatial representations. The distance of the $b$-th point along the ray is defined as:
\begin{equation}
    d_b = d_{\text{min}} + \left(b + \delta\right) \frac{d_{\text{max}} - d_{\text{min}}}{B - 1}, \quad \delta \sim \mathcal{U}(-0.5, 0.5). 
\end{equation}
Finally, we generate a dense set of $Q = N^2 \times B$ query points in 3D space. 

\subsection{Depth-guided multi-view feature sampling}

After preparing the query points, we project them onto the 2D image planes and perform depth-guided multi-view feature sampling. For a query point projected within camera $c$, we sample semantic features $f^{c}_{\text{sem}}$ from the frozen DINOv2 features~\citep{oquab2023dinov2} via bilinear interpolation, which provide objectness information. However, a fundamental challenge in multi-view feature aggregation is that a 3D query point may not be visible in some views due to occlusions or out-of-bounds projection. Therefore, we propose to use Depth-Anything-3 (DA3)~\citep{lin2025depth} to estimate multi-view consistent metric depth $\tilde{z}^*$ for guiding feature sampling. 

To align $\tilde{z}^*$ with the true depth, we use the eye position as an anchor. For each view $c$ where the head is detected, we project $X_{\text{eye}}$ into the camera coordinate system to obtain the metric eye depth $z^{c}_{\text{eye}}$. Meanwhile, we extract the estimated metric eye depth $\tilde{z}^{c*}_{\text{eye}}$ from the DA3 depth map using the head bounding box. Afterwards, we compute a view-specific scale factor in each view, take the median of these scale factors across all views to obtain a multi-view alignment scale $s_{\text{align}}$, and align the depth as $z^*$:
\begin{equation}
    s_{\text{align}} = \operatorname*{median}_{c \in [1,...,C]}(z^{c}_{\text{eye}} / \tilde{z}^{c*}_{\text{eye}}), \quad z^* = s_{\text{align}} \cdot \tilde{z}^*. 
\end{equation}

Subsequently, we compute a scene scale factor $s_{\text{scene}}$ by calculating the mean Euclidean distance between $X_{\text{eye}}$ and the optical centers of all cameras in the world coordinate system. For each query point projected into view $c$, we compute its depth difference from the corresponding DA3 surface and normalize it with $s_{\text{scene}}$: 
\begin{equation}
    \Delta z^{c} = z^{c*} - z^{c}_{\text{homo}}, \quad \Delta \bar{z}^{c} = \Delta z^{c} / s_{\text{scene}}. 
\end{equation}

We then introduce a \textit{soft depth-guided visual attention} mechanism that maps $\Delta \bar{z}^{c}$ to a continuous visibility weight:
\begin{equation}
\label{eq:vis_weight}
    w^{c} = \sigma\left(\left(\Delta \bar{z}^{c} + \epsilon\right) / \tau_{\text{occ}}\right), 
\end{equation}
where $\sigma(\cdot)$ is the sigmoid function, $\tau_{\text{occ}}$ is a temperature scalar, and $\epsilon$ is a geometric tolerance threshold. This shifted formulation preserves semantic features from geometrically visible regions ($\Delta \bar{z}^{c} \ge 0$), while smoothly suppressing those occluded by foreground surfaces ($\Delta \bar{z}^{c} < 0$). Then, we aggregate the weighted semantic features using both mean and max pooling:
\begin{equation}
    f_{\text{mean}} = \frac{\sum_{c} w^{c} f_{\text{sem}}^{c}}{\sum_{c} w^{c}}, \quad f_{\text{max}} = \max_{c \in [1,...,C]} \left(w^{c} f_{\text{sem}}^{c} \right). 
\end{equation}

Additionally, we encode $\Delta \bar{z}^{c}$ using a 1D CNN to obtain a depth embedding $f_{\text{depth}}$, capturing surface proximity and occlusion information. Finally, we concatenate $f_{\text{mean}}$, $f_{\text{max}}$, and $f_{\text{depth}}$ as the representation $f$ for each 3D query point.

\subsection{Probabilistic 3D localization}

Given the sampled features of the $Q$ query points, we perform probabilistic 3D localization. Specifically, we pass $F=\{f_1, \dots, f_Q\}$ through a 1D CNN to predict raw logit scores. Then, we apply a global softmax over all query points within the 3D gaze cone to obtain a spatial likelihood distribution $P=\{p_1, \dots, p_Q\}$. Subsequently, we estimate the 3D gaze target $X_{\text{gaze}}$ by computing the expectation over this distribution: 
\begin{equation}
    X_{\text{gaze}} = \mathbb{E}[P] = \sum_{q=1}^{Q} p_q \cdot X_q, 
\end{equation}
where $X_q$ denotes the 3D position of the $q$-th query point. 

Due to the severe noise of the pseudo 2D gaze target labels, instead of directly supervising $X_{\text{gaze}}$, we supervise the predicted distribution $P$ by generating a target distribution $P^*$. Specifically, we compute a semantic probability distribution $P^*_{\text{sem}}=\{p^*_{\text{sem},1}, \dots, p^*_{\text{sem},Q}\}$ from the pseudo 2D gaze targets and a surface probability distribution $P^*_{\text{surf}}=\{p^*_{\text{surf},1}, \dots, p^*_{\text{surf},Q}\}$ from the estimated depth, and combine them to obtain the final target distribution. 

Let $X^c_{q}$ denote the 2D projection of the $q$-th query point in view $c$, and let $X_{\text{gaze}}^{c*}$ denote the pseudo label. We map their L2 distance into a view-specific distribution using softmax and average these distributions across all views:
\begin{equation}
\label{eq:sem_distribution}
    p_{\text{sem},q}^{*} = \frac{1}{C} \sum_{c} \frac{\exp\left( -\|X^c_{q} - X_{\text{gaze}}^{c*}\|_2^2 / \tau_{\text{sem}} \right)}{\sum_{i=1}^Q \exp\left( -\|X^c_{i} - X_{\text{gaze}}^{c*}\|_2^2 / \tau_{\text{sem}} \right)}, 
\end{equation}
where $\tau_{\text{sem}}$ is a temperature parameter. Then, based on the absolute normalized metric distance $|\Delta \bar{z}^{c}_q|$ between a query point and the surface in view $c$, we further construct the surface distribution:
\begin{equation}
\label{eq:depth_distribution}
    |\Delta \bar{z}_q|=\min_{c \in [1,...,C]} |\Delta \bar{z}^{c}_q|, \quad p^{*}_{\text{surf},q} = \frac{\exp\left(-|\Delta \bar{z}_q| / \tau_{\text{surf}}\right)}{\sum_{i=1}^Q \exp\left(-|\Delta \bar{z}_i| / \tau_{\text{surf}}\right)},
\end{equation}
where $\tau_{\text{surf}}$ is a temperature parameter. 

We then use $P^*_{\text{sem}}$ to restrict $P^*_{\text{surf}}$ to semantically plausible regions. Specifically, we construct a Region-of-Interest (ROI) mask $M_{\text{ROI}}$ that selects the regions with high semantic probabilities:
\begin{equation}
\label{eq:roi_mask}
    M_{\text{ROI}, q} = \mathbb{I}\Big(p_{\text{sem},q}^{*} > \theta_{\text{ROI}} \max(P_{\text{sem}}^{*})\Big), 
\end{equation}
where $\mathbb{I}$ is the indicator function and $\theta_{\text{ROI}}$ is a tolerance threshold. We then use the ROI mask to generate the final target distribution $P^*=\{p^*_{1}, \dots, p^*_{Q}\}$:
\begin{equation}
    p^*_q = \frac{p^{*}_{\text{surf},q} \cdot M_{\text{ROI}, q}}{\sum_{i=1}^Q \left( p^{*}_{\text{surf},i} \cdot M_{\text{ROI}, i} \right)}. 
\end{equation}

Subsequently, we use Kullback-Leibler (KL) divergence loss to train the model:
\begin{equation}
    \mathcal{L}_{\text{KL}} = D_{\text{KL}}(P^*|| P) = \sum_{q=1}^Q p^{*}_q \log\left(p^{*}_q / p_q\right)
\end{equation}

To further predict whether the gaze target lies within each view, we introduce a differentiable geometric gate based on $X_{\text{gaze}}$. Let $(x_{\text{gaze}}, y_{\text{gaze}})$ denote its normalized 2D projection, and $z_{\text{gaze}}$ denote its depth in a given camera. We compute a continuous geometric probability $p_{\text{geom}}$ using sigmoid functions $\sigma(\cdot)$:
\begin{equation}
\label{eq:loss_geom}
\begin{aligned}
    p_{\text{geom}} = \sigma(\gamma_x x_{\text{gaze}}) \sigma(\gamma_x (1 - x_{\text{gaze}})) \sigma(\gamma_y y_{\text{gaze}})\\ \sigma(\gamma_y (1 - y_{\text{gaze}})) \sigma(\gamma_z z_{\text{gaze}}), 
\end{aligned}
\end{equation}
where $\gamma_x$,  $\gamma_y$, and $\gamma_z$ are scaling factors. 

Additionally, we aggregate the semantic and depth features using $P$ and use a two-layer MLP to predict a learned in-or-out probability $p_{\text{learn}}$. In the end, we multiply $p_{\text{geom}}$ and $p_{\text{learn}}$ to obtain the final in-or-out probability $p_{\text{in/out}}$:
\begin{equation}
    p_{\text{in/out}} = p_{\text{geom}} * p_{\text{learn}}.
\end{equation}
Therefore, when the 3D gaze target projects outside the valid image region, $p_{\text{geom}}$ decreases accordingly, enforcing geometric consistency in the final in-or-out prediction.

\begin{table*}
\begin{center}
\scalebox{0.8}{
\begin{tabular}{l|cccccc}
\toprule
Method & 2D Dist.* $\downarrow$ & AP* $\uparrow$ & 2D Dist. $\downarrow$ & AP $\uparrow$ & 3D Dist. [cm] $\downarrow$ & Ang. $\downarrow$ \\
\midrule
GaT~\citep{vuillecard2025enhancing} & - & - & - & - & - & 24.34 \\
GaT (m.v.) & - & - & - & - & - & 13.28 \\
Gaze-LLE~\citep{ryan2025gaze} & 0.200 & 0.831 & - & - & - & - \\
Gaze-LLE w GaT & 0.167 & 0.888 & - & - & - & - \\
Gaze-LLE w GaT (multi-view) & 0.165 & 0.890 & - & - & - & - \\
Gaze-LLE* & 0.142 & 0.918 & - & - & - & - \\
Gaze-LLE* w GaT (multi-view) & 0.137 & 0.919 & - & - & - & - \\
MV-GTE*~\citep{miao2025multi} & 0.119 & 0.907 & - & - & 97.86 & 18.69 \\
\midrule
Self-MVGTE (ours) & \textbf{0.093} & \textbf{0.929} & \textbf{0.108} & \textbf{0.910} & \textbf{74.09} & \textbf{13.08} \\
\bottomrule
\end{tabular}
}
\end{center}
\caption{Results on the MVGT dataset~\citep{miao2025multi}. Methods marked with * are trained on MVGT using ground-truth labels. Metrics marked with * are evaluated only on views where the head is visible. Self-MVGTE can project its estimated 3D gaze target onto views where the head is not visible, whereas monocular methods and MV-GTE~\citep{miao2025multi} require a visible head for 2D prediction. 3D Dist. is measured in centimeters. The 3D Dist. of MV-GTE* is obtained by triangulating its per-view 2D predictions.}
\label{tab:comparison_sota}
\end{table*}

During training, we compute $\mathcal{L}_{\text{in/out}}$ using binary cross-entropy between $p_{\text{in/out}}$ and the pseudo labels. Finally, we compute the multitask loss, where $\lambda$ is a predefined weight: 
\begin{equation}
\label{eq:loss}
    \mathcal{L} = \lambda\mathcal{L}_{\text{KL}} + \mathcal{L}_{\text{in/out}}.
\end{equation}

\section{Experiments}
\label{sec:experiments}

\noindent\textbf{Datasets: } We conduct experiments on the MVGT dataset~\citep{miao2025multi}. To the best of our knowledge, MVGT is currently the only benchmark for multi-view gaze target estimation. It contains four real-world scenes (``commons'', ``kitchen'', ``lab'', and ``shop'') recorded by six cameras, with varying camera poses across scenes. To simulate the practical scenario of applying a trained model to an unseen scene, we follow previous work~\citep{miao2025multi} and perform leave-one-scene-out cross-validation. However, in the ``shop'' scene, a person's head may only be visible in one view, while our approach requires detections in at least two views for triangulation. Therefore, we remove such cases (around 25\% of ``shop'') from both training and testing for all comparison experiments. 

\noindent\textbf{Evaluation metrics: } We evaluate our approach using both 2D and 3D metrics. For 2D metrics, we report the normalized L2 distance (2D Dist.) between the predicted and ground-truth 2D gaze target coordinates, and average precision (AP) for in-or-out classification. For 3D metrics, we report the L2 distance (3D Dist.) between the predicted and ground-truth 3D gaze target coordinates, and angular error (Ang.) between the predicted and ground-truth 3D gaze directions. 

\noindent\textbf{Implementation details: } We use a frozen DINOv2 ViT-L backbone for both monocular and multi-view models. The input image size $H \times W$ is set to $448 \times 448$, $H_D \times W_D$ to $32 \times 32$, and $d_{\text{model}}$ to 256. Following previous work~\citep{ryan2025gaze}, we train the monocular model on GazeFollow for 15 epochs using the Adam optimizer with an initial learning rate of 1e-3 and batch size 60, and then finetune it on both GazeFollow and VideoAttentionTarget for 8 epochs with a learning rate of 1e-5. $\theta_{\text{err}}$ is set to 30 degrees. For the 3D gaze cone, $\theta_{\text{ang}}$ is set to 12 degrees, $N$ is set to 5, and $B$ is set to 48. In Eq.~\ref{eq:vis_weight}, $\epsilon$ is set to 0.05 and $\tau_{\text{occ}}$ is set to 0.15. In Eq.~\ref{eq:sem_distribution}, $\tau_{\text{sem}}$ is set to 0.01. In Eq.~\ref{eq:depth_distribution}, $\tau_{\text{surf}}$ is set to 0.05. In Eq.~\ref{eq:roi_mask}, $\theta_{\text{ROI}}$ is set to 0.2. In Eq.~\ref{eq:loss_geom}, $\gamma_x$,  $\gamma_y$, and $\gamma_z$ are all set to 10.0. In Eq.~\ref{eq:loss}, $\lambda$ is set to 10.0.

\subsection{Comparison with the state-of-the-art}

\noindent\textbf{3D gaze estimation: } We use an off-the-shelf GaT model~\citep{vuillecard2025enhancing} trained on Gaze360~\citep{kellnhofer2019gaze360} and GazeFollow~\citep{recasens2015they} to estimate the 3D gaze directions on the MVGT dataset. We also average the monocular predictions across views to obtain a multi-view prediction. As shown in Table~\ref{tab:comparison_sota}, the averaging strategy reduces the angular error by over 10 degrees. We use the multi-view predictions as geometric priors for 3D gaze target estimation. 

\noindent\textbf{Monocular gaze target estimation: } We evaluate the performance of Gaze-LLE with 3D priors on the unseen MVGT dataset. We compare our approach with: (1) the original Gaze-LLE trained on GazeFollow~\citep{recasens2015they} and VideoAttentionTarget~\citep{chong2020detecting}; (2) the original Gaze-LLE fine-tuned on MVGT; (3) Gaze-LLE with 3D priors fine-tuned on MVGT. As shown in Table~\ref{tab:comparison_sota}, adding 3D priors slightly improves the fine-tuned Gaze-LLE, while significantly improving its cross-dataset performance on the unseen data (0.200 \textit{v.s} 0.165 2D Dist. and 0.831 \textit{v.s} 0.890 AP). We also observe that using monocular or multi-view gaze vectors as 3D priors leads to similar performance on MVGT. These results demonstrate the benefit of explicitly encoding 3D gaze directions for vision foundation model-based gaze target estimation in unseen scenes, even when the 3D gaze estimation is not that accurate. 

\noindent\textbf{Multi-view gaze target estimation: } Using the predictions of ``Gaze-LLE w GaT (m.v.)'' as pseudo labels, we train Self-MVGTE and compare it with state-of-the-art monocular and multi-view methods. However, for 2D evaluation, Self-MVGTE can project the estimated 3D gaze target onto a view even when the head is not visible in that view. In contrast, monocular methods and MV-GTE~\citep{miao2025multi} require a visible head to make a 2D prediction. Therefore, for a fair comparison, we additionally report 2D Dist* and AP* only on views where the head is visible. 

Compared with Gaze-LLE and MV-GTE trained with ground-truth annotations, Self-MVGTE achieves better performance across all metrics, showing the effectiveness of the self-supervised 3D learning framework. Although MV-GTE predicts per-view 2D gaze targets, it internally predicts a 3D gaze vector, which yields a much larger angular error. We further triangulate the per-view predictions of MV-GTE to obtain a 3D gaze target, which results in a much larger 3D Dist., highlighting the limitation of recovering 3D gaze targets from independent 2D predictions. We also observe that the fully-supervised Gaze-LLE obtains a higher AP than the fully-supervised MV-GTE. This indicates an important distinction between 2D semantic classification and 3D geometric projection, as MV-GTE is structurally affected by multi-view geometry. Nevertheless, Self-MVGTE still significantly outperforms both Gaze-LLE and the pseudo labels used for training, proving the effectiveness of our geometry-based framework. 

\begin{figure*}[t!]
    \centering
    \includegraphics[width=0.9\textwidth]{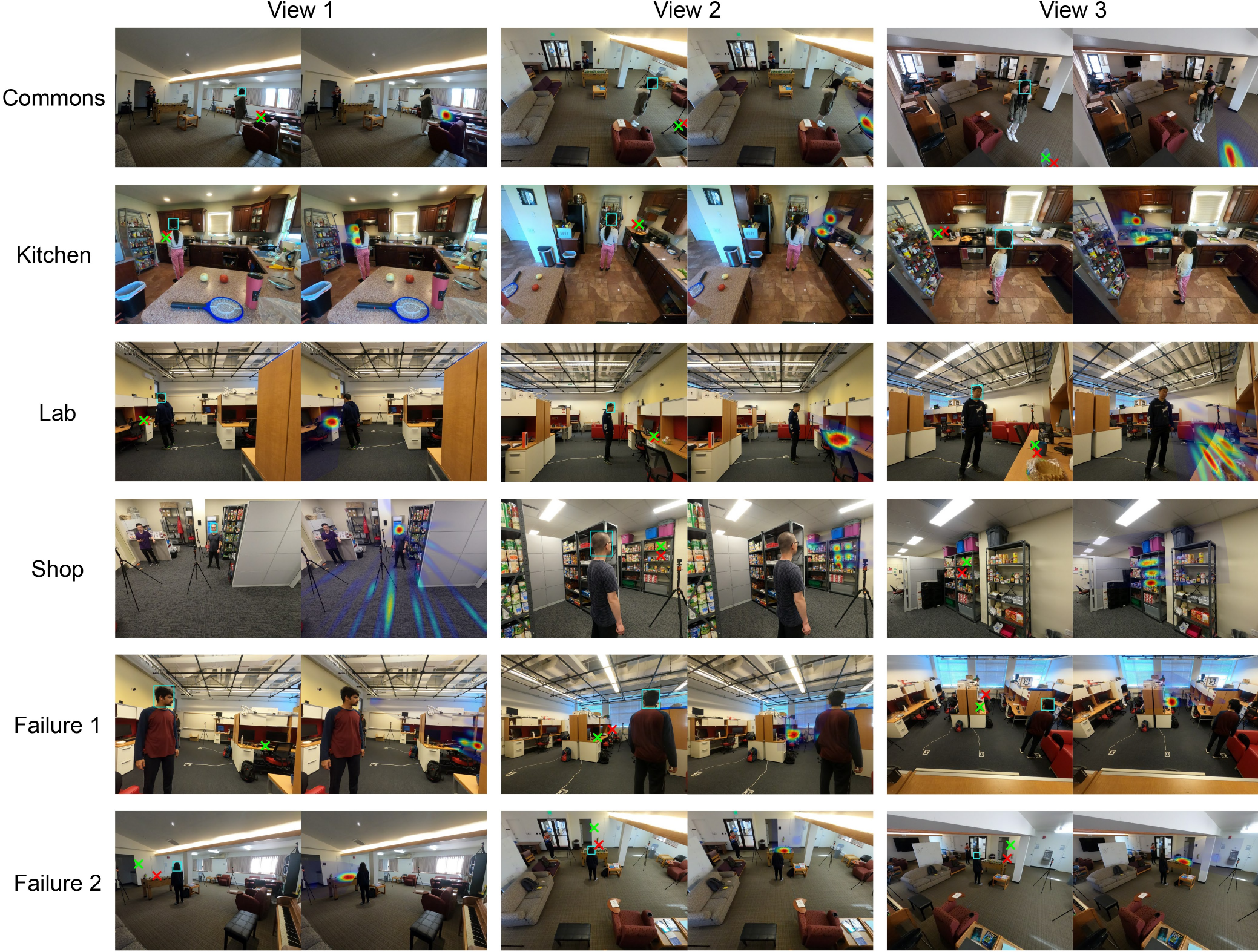} 
    \caption{Qualitative results on MVGT~\citep{miao2025multi} with visualized gaze cones. Ground-truth and predictions are green and red, respectively. Failure case 1 is caused by inaccurate gaze estimation, where the target is outside the gaze cone (see view 3); failure case 2 is caused by semantic bias, where the person is actually looking at the white wall instead of the table. }
    \label{fig:results}
\end{figure*}

\begin{table}
\begin{center}
\scalebox{0.8}{
\begin{tabular}{c|cccc}
\toprule
Multi-view fusion & 2D Dist. $\downarrow$ & AP $\uparrow$ & 3D Dist. $\downarrow$ & Ang. $\downarrow$ \\
\midrule
Training-free geometry & 0.158 & 0.843 & 116.87 & 23.92 \\
Learned triangulation & 0.169 & 0.837 & 136.16 & 23.02 \\
MLP fusion & 0.162 & 0.747 & 133.11 & 22.79 \\
Self-attention fusion & 0.240 & 0.710 & 173.60 & 44.66 \\
\midrule
Probabilistic ray marching & \textbf{0.108} & \textbf{0.910} & \textbf{74.09} & \textbf{13.08} \\
\bottomrule
\end{tabular}
}
\end{center}
\caption{Ablation study of different multi-view fusion strategies. }
\label{tab:abl_fusion}
\end{table}

\begin{table}
\begin{center}
\scalebox{0.8}{
\begin{tabular}{c|cccc}
\toprule
3D gaze estimation (Ang.) & 2D Dist. $\downarrow$ & AP $\uparrow$ & 3D Dist. $\downarrow$ & Ang. $\downarrow$ \\
\midrule
Baseline (13.28) & 0.108 & 0.910 & 74.09 & 13.08 \\
Pseudo GT (11.82) & 0.100 & 0.908 & 67.48 & 11.30 \\
GT (0.0) & \textbf{0.068} & \textbf{0.938} & \textbf{45.30} & \textbf{3.88} \\
\bottomrule
\end{tabular}
}
\end{center}
\caption{An oracle analysis of using more accurate 3D gaze estimation. }
\label{tab:abl_oracle}
\end{table}

\begin{table}
\begin{center}
\scalebox{0.8}{
\begin{tabular}{c|cccc}
\toprule
Localization loss & 2D Dist. $\downarrow$ & AP $\uparrow$ & 3D Dist. [cm] $\downarrow$ & Ang. $\downarrow$ \\
\midrule
L2 loss & 0.142 & 0.878 & 116.05 & 13.63 \\
KL loss & \textbf{0.108} & \textbf{0.910} & \textbf{74.09} & 13.08 \\
L2 \& KL loss & 0.109 & 0.903 & 76.97 & \textbf{13.06} \\
\bottomrule
\end{tabular}
}
\end{center}
\caption{Ablation study of different localization loss functions. }
\label{tab:abl_l2}
\end{table}

\begin{table}
\begin{center}
\scalebox{0.8}{
\begin{tabular}{c|cccc}
\toprule
Pseudo labels & 2D Dist. $\downarrow$ & AP $\uparrow$ & 3D Dist. $\downarrow$ & Ang. $\downarrow$ \\
\midrule
Gaze-LLE & 0.112 & 0.896 & 76.26 & \textbf{13.00} \\
Gaze-LLE w GaT (m.v.) & \textbf{0.108} & \textbf{0.910} & \textbf{74.09} & 13.08 \\
\bottomrule
\end{tabular}
}
\end{center}
\caption{Ablation study of different pseudo gaze target labels. }
\label{tab:abl_pseudo}
\end{table}

\begin{table}
\begin{center}
\scalebox{0.8}{
\begin{tabular}{c|cccc}
\toprule
$\theta_{\text{ang}}$ and grid size & 2D Dist. $\downarrow$ & AP $\uparrow$ & 3D Dist. [cm] $\downarrow$ & Ang. $\downarrow$ \\
\midrule
6 degrees, $3 \times 3$ & 0.111 & 0.902 & 78.84 & 12.96 \\
6 degrees, $5 \times 5$ & 0.111 & 0.904 & 78.14 & 13.01 \\
6 degrees, $7 \times 7$ & 0.111 & 0.899 & 78.27 & 13.05 \\
\midrule
9 degrees, $3 \times 3$ & 0.109 & 0.908 & 76.51 & 13.10 \\
9 degrees, $5 \times 5$ & 0.110 & 0.903 & 76.97 & \textbf{12.88} \\
9 degrees, $7 \times 7$ & 0.109 & 0.906 & 75.38 & 12.89 \\
\midrule
12 degrees, $3 \times 3$ & 0.111 & 0.906 & 77.87 & 13.30 \\
12 degrees, $5 \times 5$ & \textbf{0.108} & \textbf{0.910} & 74.09 & 13.08 \\
12 degrees, $7 \times 7$ & 0.109 & 0.907 & \textbf{73.71} & 13.07 \\
\midrule
15 degrees, $5 \times 5$ & 0.111 & 0.907 & 76.58 & 13.27 \\
15 degrees, $7 \times 7$ & 0.110 & 0.907 & 74.10 & 13.11 \\
15 degrees, $9 \times 9$ & 0.110 & 0.904 & 75.45 & 13.11 \\
\midrule
18 degrees, $5 \times 5$ & 0.114 & 0.907 & 77.81 & 13.76 \\
18 degrees, $7 \times 7$ & 0.113 & 0.907 & 77.38 & 13.59 \\
18 degrees, $9 \times 9$ & 0.111 & 0.907 & 76.58 & 13.52 \\
\bottomrule
\end{tabular}
}
\end{center}
\caption{Ablation study of the spread angle $\theta_{\text{ang}}$ and grid size $N$. }
\label{tab:abl_angle}
\end{table}

\subsection{Ablation study}

\noindent\textbf{Multi-view fusion strategies: } We compare our probabilistic ray marching framework with the other multi-view fusion strategies, including: (1) training-free geometry based on pure triangulation; (2) learned triangulation, which predicts per-view confidence weights for weighted triangulation; (3) MLP-based fusion and self-attention-based fusion, which refine per-view 2D gaze targets and predict confidence weights for triangulation. As shown in Table~\ref{tab:abl_fusion}, our approach achieves the best performance, while the other strategies are more sensitive to noisy pseudo labels. 

\noindent\textbf{Oracle 3D gaze estimation: } We evaluate whether Self-MVGTE benefits from more accurate 3D gaze estimation. Specifically, we compare two oracle settings: (1) following~\citep{vuillecard2025enhancing}, we geometrically rotate the original predicted 3D gaze vector so that its 2D projection is aligned with the ground-truth 2D gaze label, which reduces the angular error by 1.46 degrees; (2) we directly use ground-truth 3D gaze vectors. As shown in Table~\ref{tab:abl_oracle}, both settings substantially improve the performance. These results show that Self-MVGTE can benefit from future improvements in 3D gaze estimation. 

\noindent\textbf{L2 loss \textit{v.s} KL loss: } We evaluate the effectiveness of probabilistic learning by comparing two localization objectives: (1) L2 loss that directly supervises the predicted gaze target using pseudo labels; (2) KL loss that supervises the predicted spatial distribution using target distributions converted from the pseudo labels. As shown in Table~\ref{tab:abl_l2}, using the KL loss solely achieves the best performance. 

\noindent\textbf{Pseudo gaze target labels: } We evaluate the influence of pseudo gaze target label quality by training Self-MVGTE using pseudo labels generated by the original Gaze-LLE. As shown in Table~\ref{tab:abl_pseudo}, the performance decreases slightly. 

\noindent\textbf{3D gaze cone: } We evaluate how the spread angle $\theta_{\text{ang}}$ and grid size $N$ of the 3D gaze cone affect performance on the MVGT dataset. As shown in Table~\ref{tab:abl_angle}, when $\theta_{\text{ang}}$ is around 12 degrees and $N$ is set to 5 or 7, we obtain the best performance. Nevertheless, the performance is relatively insensitive to small variations in these two hyperparameters.

\noindent\textbf{Depth: } We evaluate the effectiveness of using the estimated depth from DA3. As shown in Table~\ref{tab:abl_depth}, adding depth significantly improves the overall performance. 

\noindent\textbf{Number of views: } We evaluate the performance of Self-MVGTE using different numbers of camera views. During both training and testing, we vary the number of views by random selection, while prioritizing views with visible heads for triangulation. As shown in Table~\ref{tab:abl_view_num}, the 3D Dist. increases noticeably when fewer than four views are used.

\begin{table}
\begin{center}
\scalebox{0.8}{
\begin{tabular}{c|cccc}
\toprule
Using depth & 2D Dist. $\downarrow$ & AP $\uparrow$ & 3D Dist. [cm] $\downarrow$ & Ang. $\downarrow$ \\
\midrule
\xmark & 0.125 & 0.892 & 99.51 & 13.55 \\
\checkmark & \textbf{0.108} & \textbf{0.910} & \textbf{74.09} & \textbf{13.08} \\
\bottomrule
\end{tabular}
}
\end{center}
\caption{Ablation study of using depth. }
\label{tab:abl_depth}
\end{table}

\begin{table}
\begin{center}
\scalebox{0.8}{
\begin{tabular}{c|cccc}
\toprule
Number of views & 2D Dist. $\downarrow$ & AP $\uparrow$ & 3D Dist. [cm] $\downarrow$ & Ang. $\downarrow$ \\
\midrule
6 & 0.108 & \textbf{0.910} & \textbf{74.09} & 13.08 \\
5 & 0.104 & 0.896 & 74.75 & 12.84 \\
4 & \textbf{0.098} & 0.901 & 74.35 & \textbf{12.69} \\
3 & 0.108 & 0.905 & 82.81 & 14.67 \\
2 & 0.117 & 0.894 & 102.03 & 17.13 \\
\bottomrule
\end{tabular}
}
\end{center}
\caption{Ablation study of using different numbers of views. }
\label{tab:abl_view_num}
\end{table}

\begin{table}
\begin{center}
\scalebox{0.8}{
\begin{tabular}{c|ccc}
\toprule
Metric & Only $l_{\text{learn}}$ & Only $l_{\text{geom}}$ & Using $l_{\text{geom}} * l_{\text{learn}}$ \\
\midrule
AP $\uparrow$  & 0.890 & 0.897 & \textbf{0.910} \\
\bottomrule
\end{tabular}
}
\end{center}
\caption{Ablation study of using geometric gate for in-or-out prediction. }
\label{tab:abl_geom}
\end{table}

\noindent\textbf{Geometric gate: } We evaluate the significance of the differentiable geometric gate based on 3D-to-2D projection for in-or-out prediction. As shown in Table~\ref{tab:abl_geom}, using only $l_{\text{learn}}$ performs the worst, which is the strategy of traditional 2D gaze target estimation; using only $l_{\text{geom}}$, which contains no learnable parameters, performs slightly better; using both terms performs the best. This demonstrates the benefit of leveraging 3D gaze target localization for in-or-out prediction.

\subsection{Qualitative analysis and limitations}

We present qualitative results in Figure~\ref{fig:results}, including visualization of the predicted spatial distributions within the gaze cone. In the ``kitchen'' scene, the highest likelihoods are concentrated around the cupboard and the table. In view 1 of the  ``shop'' scene, no semantic objects fall within the gaze cone. Though our model performs well in general, we identify two kinds of failure cases: (1) the ground-truth gaze target falls outside the 3D gaze cone due to inaccurate 3D gaze estimation; (2) the model exhibits a bias toward semantically rich regions, incorrectly localizing the gaze target at the table while the person is actually looking at the white wall. 

\section{Conclusion}
\label{sec:conclusion}

We present Self-MVGTE, the first approach for multi-view 3D gaze target estimation. Unlike existing gaze target estimation methods that localize independent 2D gaze targets in each view, Self-MVGTE directly localizes the gaze target in 3D space. Moreover, we address this new task in a self-supervised manner without requiring 2D or 3D ground-truth annotations from the target scene. Using estimated 3D gaze vectors as geometric priors, we first improve the generalization of a state-of-the-art monocular gaze target estimation model to unseen scenes. We then introduce a probabilistic ray marching framework that constructs a constrained 3D gaze cone around the estimated gaze direction and performs depth-guided multi-view feature sampling using DINOv2 and Depth-Anything-3. Within this cone, the model estimates a spatial likelihood distribution of the gaze target and learns from noisy pseudo labels through probabilistic learning. Experiments on the MVGT dataset demonstrate that Self-MVGTE surpasses existing fully-supervised baselines, proving its effectiveness.

\section{Acknowledgments}
This work was supported by French state funds managed within the Plan Investissements d’Avenir by the ANR under references ANR-22-FAI1-0001 (project DAIOR), ANR-10-IAHU-02 (IHU Strasbourg) and ANR-23-IACL-0004 (ENACT AI Cluster). This work was also granted access to the servers/HPC resources managed by CAMMA, IHU Strasbourg, Unistra Mesocentre, and GENCI-IDRIS [Grant 2021-AD011011638R3].

\bibliographystyle{model2-names.bst}
\bibliography{main}

\end{document}